\documentclass[lettersize,journal]{IEEEtran}
\usepackage{amsmath,amsfonts}
\usepackage{algorithmic}
\usepackage{algorithm}
\usepackage{array}
\usepackage[caption=false,font=normalsize,labelfont=sf,textfont=sf]{subfig}
\usepackage{textcomp}
\usepackage{stfloats}
\usepackage{url}
\usepackage{verbatim}
\usepackage{graphicx}
\usepackage{cite}
\usepackage{hyperref}
\usepackage{booktabs}
\usepackage{caption}
\begin{document}

\title{Shape-IoU: More Accurate Metric considering Bounding Box Shape and Scale}

\author{Hao Zhang, Shuaijie Zhang}

\maketitle

\begin{abstract}
As an important component of the detector localization branch, bounding box regression loss plays a significant role in object detection tasks. The existing bounding box regression methods usually consider the geometric relationship between the GT box and the predicted box, and calculate the loss by using the relative position and shape of the bounding boxes, while ignoring the influence of inherent properties such as the shape and scale of the bounding boxes on bounding box regression. In order to make up for the shortcomings of existing research, this article proposes a bounding box regression method that focuses on the shape and scale of the bounding box itself. Firstly, we analyzed the regression characteristics of the bounding boxes and found that the shape and scale factors of the bounding boxes themselves will have an impact on the regression results. Based on the above conclusions, we propose the Shape IoU method, which can calculate the loss by focusing on the shape and scale of the bounding box itself, thereby making the bounding box regression more accurate. Finally, we validated our method through a large number of comparative experiments, which showed that our method can effectively improve detection performance and outperform existing methods, achieving state-of-the-art performance in different detection tasks.Code is available at \url{https://github.com/malagoutou/Shape-IoU}.

\end{abstract}


\section{Introduction}
\IEEEPARstart{O}{bject} detection is one of the basic tasks of computer vision, which aims to locate and recognize the object in an image. 
They can be categorized into Anchor-based and Anchor-free methods according to whether they generate an Anchor or not.
Anchor based algorithms include FasterR-CNN\cite{ref1}, YOLO (You Only Look Once) series\cite{ref2},SSD (Single Shot MultiBox Detector)\cite{ref3} and RetinaNet\cite{ref4}.Anchor free detection algorithms include CornerNet\cite{ref5},CenterNet\cite{ref6} and FCOS (Fully Convolutional One StageObject Detection)\cite{ref7}. In these detectors, the bounding box regression loss function plays an irreplaceable role as an important component of the localization branch.
\par The most commonly used methods in the field of object detection are IoU\cite{ref8}, GIoU\cite{ref9}, CIoU\cite{ref10}, SIoU\cite{ref11}, etc. IoU\cite{ref8} as the most widely used loss function in the field of object detection, which has the advantage of more accurately describing the degree of match between the prediction box and the GT box. Its deficiency mainly lies in the inability to accurately describe the positional relationship between the prediction box and the GT box when the overlap between the two boxes is 0. GIoU\cite{ref9} provides a specific improvement to this deficiency by introducing the minimum enclosing box. CIoU \cite{ref10}further improves the detection accuracy by adding a shape loss term on the basis of considering the minimization of the normalized distance between the centroid of the prediction box and the GT box. In the work of SIoU\cite{ref11}, it is proposed to add the angular size of the line connecting the center point of the prediction box and the GT box as a new loss term to be considered, so as to judge the degree of matching between the prediction box and the GT box more accurately through the change of the angle.
In conclusion, the previous methods of bounding box regression mainly achieve  regression more accurate by adding new geometric constraints on top of IoU\cite{ref8}. The above methods considered the influence of the distance, shape and angle of the GT box and the Anchor box on the bounding box regression, but neglected the fact that the shape and scale of the bounding box itself will also have influences on the bounding box regression. In order to further improve the accuracy of regression, we analyze the influence of the shape and angle of the bounding box itself and propose a new generation of bounding regression loss : Shape-IoU.
\begin{figure}[!t]
	\centering
	\includegraphics[width=\linewidth]{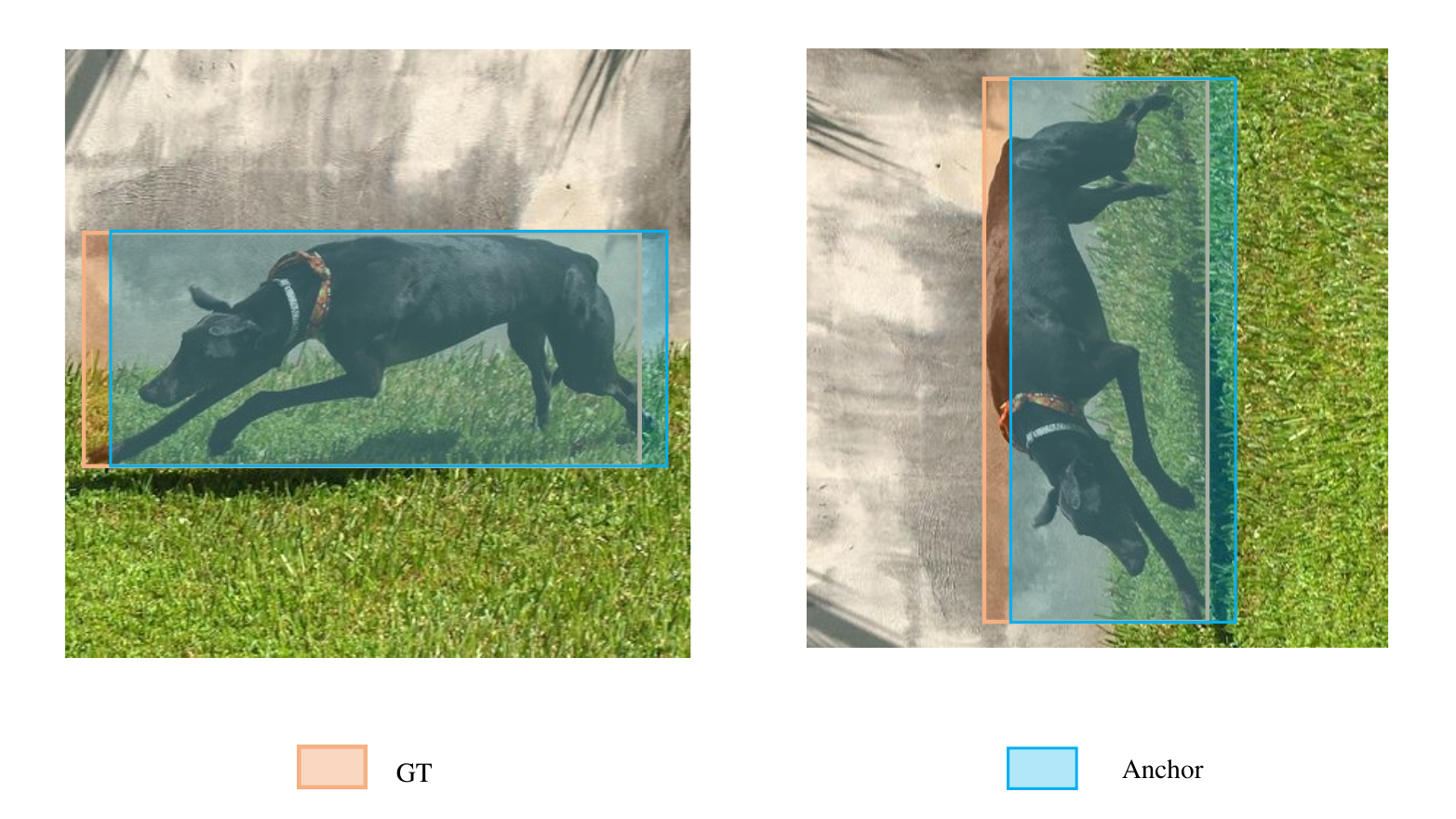}
	\caption{The regression samples in the left of the figure and the right of the figure differ only in the shape of the bounding box, and the deviation of the two regression samples is the same corresponding to the direction of the long side and the direction of the short side of the GT box, respectively, and the difference in the regression effect caused by the shape factor of the GT box can be seen from the figure. The regression result of the left figure is better than that of the right figure.}
	\label{fig_1} 
\end{figure}
\par The main contributions of this article are as follows:
\par$\bullet$We analyze the characteristics of bounding box regression and conclude that in the process of bounding box regression, the shape and scale factors of the bounding box regression samples themselves will have impacts on the results of regression.
\par$\bullet$Based on the existing loss function of bounding box regression, considering the influence of the shape and scale of the bounding box regression samples themselves on the bounding box regression, we propose the shape-IoU loss function, and for the task of tiny target detection we propose the shape-dot-distance and shape-nwd loss.
\par$\bullet$We conduct a series of comparison experiments on different detection tasks using state-of-the-art one-stage detectors, and the experimental results prove that the detection effect of the method in this paper is better than the existing methods to achieve sota.
\begin{figure*}[!htbp]
	\subfloat[]{	
		\centering
		\includegraphics[width=3.5in]{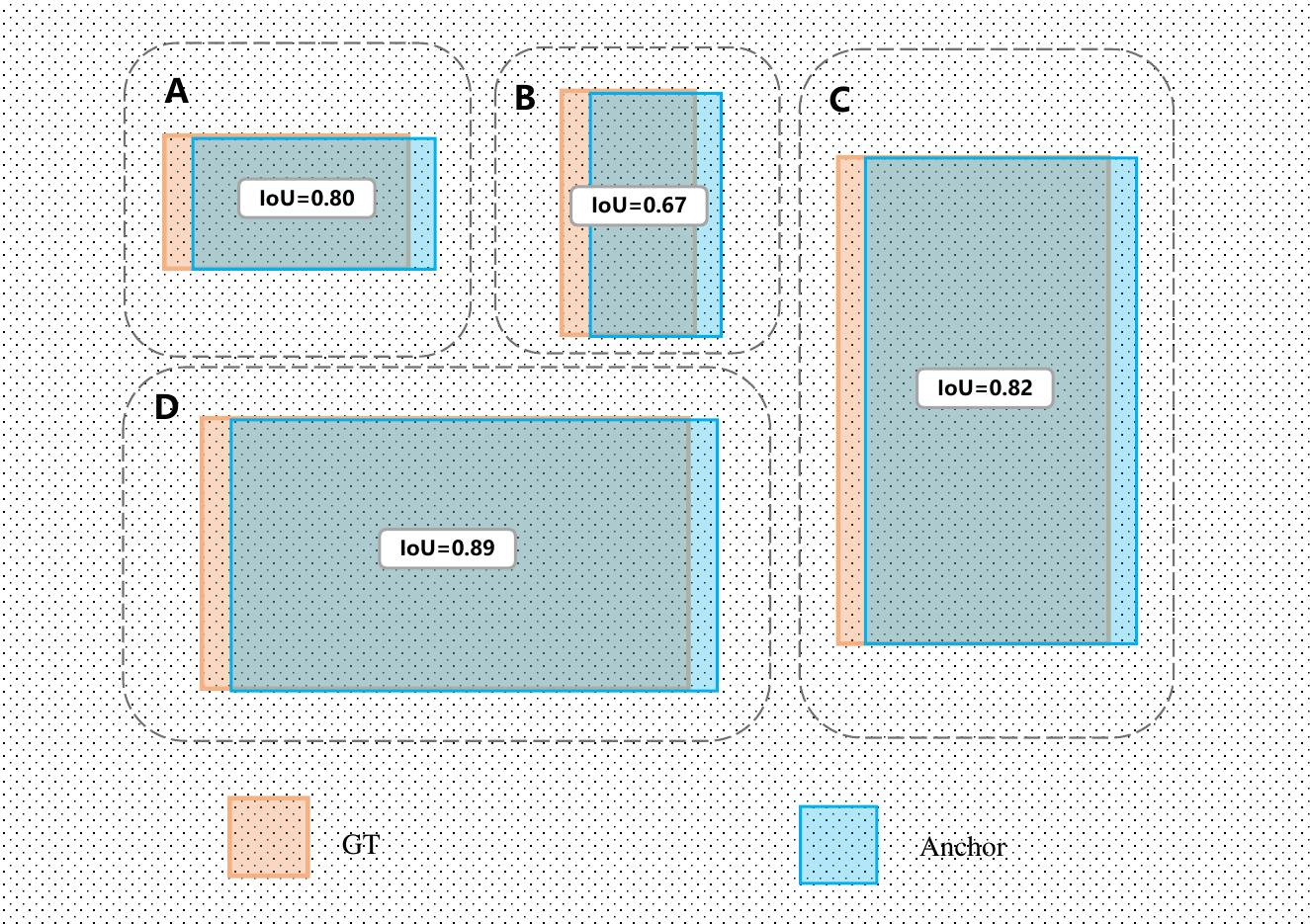}}
	\hspace{0.2cm}
	\subfloat[]{
		\centering
		\includegraphics[width=3.5in]{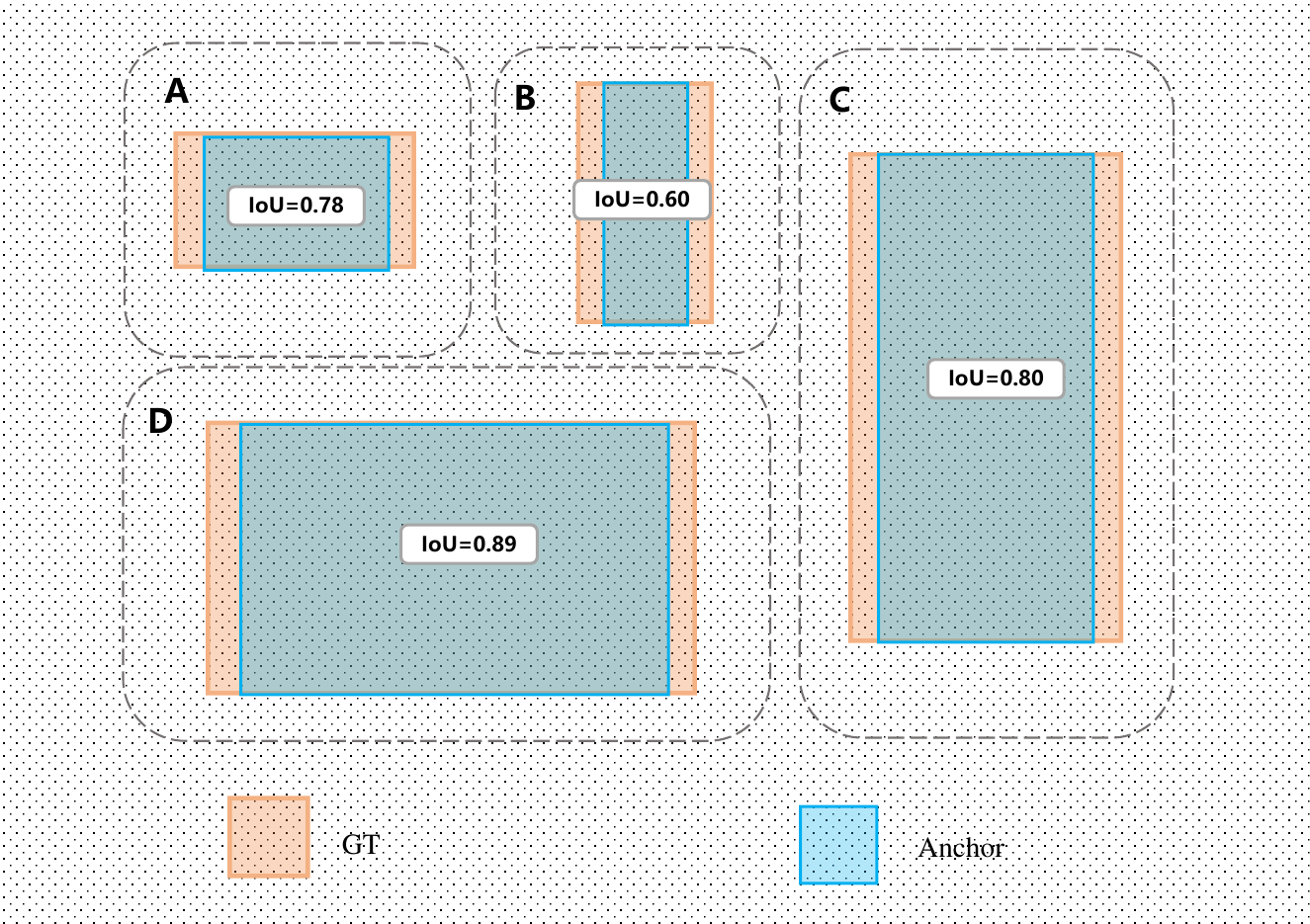}
	}
	\caption{}
	\label{fig_2}
\end{figure*}
\section{Related work}
\subsection{IoU-based Metric in Object Detection}
In recent years with the development of detectors, the bounding box regression loss has been rapidly developed. Initially IoU was proposed to be used for evaluating the state of bounding box regression, and methods such as GIoU\cite{ref9}, DIoU\cite{ref10}, CIoU\cite{ref10}, EIoU\cite{ref12}, and SIoU\cite{ref11} are continuously updated by adding different constraints on basis of IoU to achieve better detection.
\subsubsection{IoU Metric}
\par The IoU (Intersection over Union)\cite{ref8}, which is the most popular target detection evaluation criterion, is defined as follows:
\begin{equation} 
	IoU=\displaystyle\frac{\left\vert B\cap B^{gt} \right\vert}{\left\vert B\cup B^{gt} \right\vert}
\end{equation}
where $B$ and $B^{gt}$  represent the predicted box and the GT box, respectively.
\subsubsection{GIoU Metric}
In order to solve the gradient vanishing problem of IoU loss due to no overlap between GT box and Anchor box in the bounding box regression, GIoU (Generalized Intersection over Union)\cite{ref9} is proposed. Its definition is as follows:
	\begin{equation} 
	GIoU=IoU-\displaystyle\frac{\left\vert C-B\cap B^{gt} \right\vert}{\left\vert C \right\vert}
\end{equation}
 where $C$ represents the smallest enclosing box between the GT box and the Anchor box.
 \subsubsection{DIoU Metric}
 Compared with GIoU, DIoU\cite{ref10} considers the distance constraints between the bounding boxes, and by adding the centroid normalized distance loss term on the basis of IoU, its regression results are more accurate. It is defined as follows:
 	\begin{equation}
 	DIoU=IoU-\displaystyle\frac{\rho^{2}(b,b^{gt})}{c^{2}}
 \end{equation}
 Where $b$ and $b^{gt}$  are the center points of anchor box and GT box respectively, $\rho\left(\cdot\right) $  refers to the Euclidean distance, where $c$ is the diagonal distance of the minimum enclosing bounding box between $b$ and $b^{gt}$.
 \par CIoU\cite{ref10} further considers the shape similarity between GT and Anchor boxes by adding a new shape loss term to DIoU to reduce the difference in aspect ratio between Anchor and GT boxes. It is defined as follows:
 \begin{equation}
 	CIoU=IoU-\displaystyle\frac{\rho^{2}(b,b^{gt})}{c^{2}}-\alpha v
 \end{equation}
 	\begin{equation}
 	\alpha=\displaystyle\frac{v}{(1-IoU)+v}
 \end{equation}
 \begin{equation}
 	v=\displaystyle\frac{4}{\pi^{2}}(arctan\displaystyle\frac{w^{gt}}{h^{gt}}-arctan\displaystyle\frac{w}{h})^{2}
 \end{equation}
 	where $w^{gt}$ and $h^{gt}$ denote the width and height of GT box,$w$ and $h$ denote the width and height of anchor box.
  \subsubsection{EIoU Metric}
  EIoU\cite{ref12} redefines the shape loss based on CIoU, and further improves the detection accuracy by directly reducing the aspect difference between GT boxes and Anchor boxes. It is defined as follows:
  \begin{equation}
  	EIoU=IoU-\displaystyle\frac{\rho^{2}(b,b^{gt})}{c^{2}}-\displaystyle\frac{\rho^{2}(w,w^{gt})}{(w^{c})^{2}}-\displaystyle\frac{\rho^{2}(h,h^{gt})}{(h^{c})^{2}}
  \end{equation}
   Where $w^{c}$ and $h^{c}$ are the width and height of the minimum bounding box covering GT box and anchor box.
    \subsubsection{SIoU Metric}
    On the basis of previous research, SIoU\cite{ref11} further considers the influence of the angle between the bounding boxes on the bounding box regression, which aims to accelerate the convergence process by decreasing the angle between the anchor box and GT box which is the horizontal or vertical direction. Its definition is as follows:
    	\begin{equation}
    	SIoU = IoU - \displaystyle\frac{(\Delta+\Omega)}{2}
    \end{equation}
    \begin{equation}
    	\Lambda=sin(2sin^{-1}\displaystyle\frac{min(\left\vert x_{c}^{gt}-x_{c} \right\vert,\left\vert y_{c}^{gt}-y_{c} \right\vert)}{\sqrt{ (x_{c}^{gt}-x_{c})^{2}+(y_{c}^{gt}-y_{c})^{2}}+\in})
    \end{equation}
    	\begin{equation}
    	\Delta=\sum_{t=w,h}(1-e^{-\gamma\rho_{t}}),\gamma=2-\Lambda
    \end{equation}
    	\begin{equation}
    	\left\{
    	\begin{aligned}
    		\displaystyle\rho_{x} & = & (\frac{x_{c}-x_{c}^{gt}}{w^{c}})^{2} \\
    		\rho_{y} & = & (\frac{y_{c}-y_{c}^{gt}}{h^{c}})^{2} \\
    	\end{aligned}
    	\right.
    \end{equation}
    	\begin{equation}
    	\Omega=\sum_{t=w,h}(1-e^{-\omega_{t}})^{\theta},\theta=4
    \end{equation}
    	\begin{equation}
    	\left\{
    	\begin{aligned}
    		\displaystyle\omega_{w} & = & \frac{\left\vert w - w_{gt} \right\vert}{max(w,w_{gt})} \\
    		\displaystyle\omega_{h} & = & \frac{\left\vert h - h_{gt} \right\vert}{max(h,h_{gt})} \\
    	\end{aligned}
    	\right.
    \end{equation}

\subsection{Metric in Tiny Object Detection}
IoU-based metrics are suitable for general object detection tasks, and in the case of small object detection, Dot Distance\cite{ref13} and Normalized Wasserstein Distance (NWD)\cite{ref14} have been proposed in order to overcome their own sensitivity to with IoU values.
\subsubsection{Dot Distance}
\begin{equation}
D = \sqrt{ (x_c - {x_c}^{gt})^2 + (y_c - {y_c}^{gt})^2 }
\end{equation}
\begin{equation}
	S = \sqrt{\frac{\sum_{i=1}^{M} \sum_{j=1}^{N_i} w_{ij} \cdot h_{ij}}{\sum_{i=1}^{M} N_i
	}} 	
\end{equation}
\begin{equation}
	DotD = e^{-\frac{D}{S}}	
\end{equation}
Where D denotes the Euclidean distance between the center point of the GT box and the center point of the Anchor box, S refers to the average size of the target in the dataset. M refers to the number of images, ${N_i}$ refers to the number of labeled bounding boxes in the i-th image, and $w_{ij}$ and $h_{ij}$ stand for the width and height of the j-th border in the i-th image species, respectively.
\subsubsection{Normalized Gaussian Wasserstein Distance}
\begin{equation}
	D = \sqrt{ (x_c - {x_c}^{gt})^2 + (y_c - {y_c}^{gt})^2 + \frac{{(w - {w}^{gt})^2 + (h - {h}^{gt})^2}}{{\text{weight}^2}} }
\end{equation}
\begin{equation}
	NWD = e^{-\frac{D}{C}}
\end{equation}
where weight = 2 and C is the constant associated with the dataset.
\begin{figure}[!t]
	\centering
	\includegraphics[width=\linewidth]{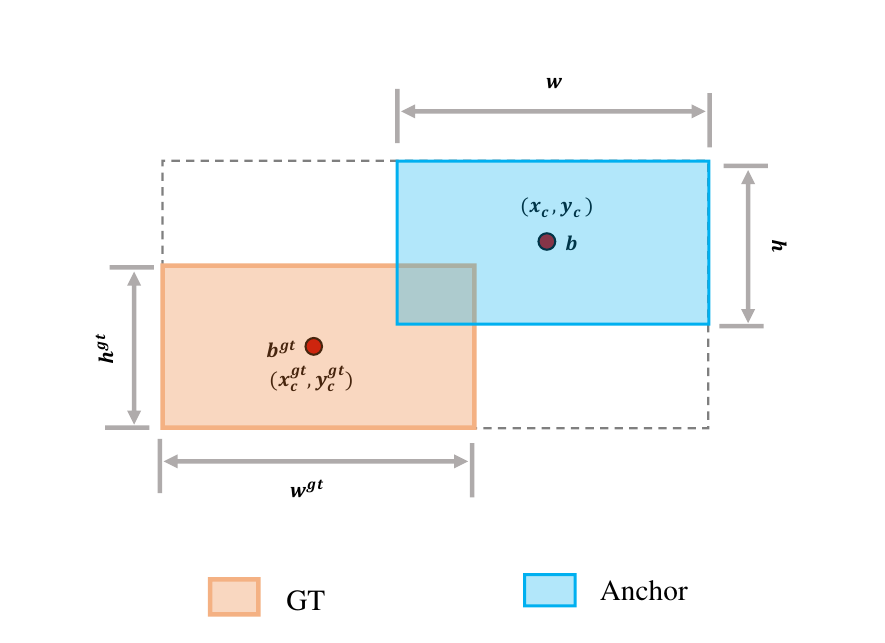}
	\caption{}
	\label{fig_3} 
\end{figure}
\section{Methods}
\subsection{Analysis of Bounding Box Regression Characteristics}
As shown in Fig.\ref{fig_2}, the scale of the GT box in bounding box regression sample A and B is the same, while the scale of the GT box in C and D is the same. The shape of the GT box in A and D is the same, while the shape of the GT box in B and C is the same. The scale of the bounding boxes in C and D is greater than the scale of the bounding boxes in A and B. The regression samples for all bounding boxes in Fig.\ref{fig_2}a have the same deviation, with a shape deviation of 0. The difference between Fig.\ref{fig_2}a and Fig.\ref{fig_2}b is that the shape-deviation of all bounding box regression samples in Fig.\ref{fig_2}b is the same, with a deviation of 0.
\par The deviation between A and B in Fig.\ref{fig_2}a is the same, but there is a difference in the IoU value.
\par The deviation between C and D in Fig.\ref{fig_2}a is the same, but there is a difference in IoU values, and compared to A and B Fig.\ref{fig_2}a, the difference in IoU values is not significant.
\par The shape-deviation of A and B in Fig.\ref{fig_2}b is the same, but there is a difference in the IoU value.
\par The shape-deviation of C and D in Fig.\ref{fig_2}b is the same, but there is a difference in IoU values, and compared to A and B in Fig.\ref{fig_2}a, the difference in IoU values is not significant.
\par The reason for the difference in IoU value between A and B in Fig.\ref{fig_2}a is that their GT boxes have different shapes, and the deviation direction corresponds to their long and short edge directions, respectively. For A, the deviation along the long edge direction of their GT boxes has a smaller impact on their IoU value, while for B, the deviation in the short edge direction has a greater impact on their IoU value. Compared to large scale bounding boxes, smaller scale bounding boxes are more sensitive to changes in IoU value, and the shape of GT boxes has a more significant impact on the IoU value of smaller scale bounding boxes. Because A and B are smaller in scale than C and D, the change in IoU value is more significant when the shape and deviation are the same. Similarly, in Fig.\ref{fig_2}b, analyzing bounding box regression from the perspective of shape deviation reveals that the shape of the GT box in the regression sample will affect its IoU value during the regression process.
\par Based on the above analysis, the following conclusions can be drawn:
\par$\bullet$Assuming that the GT box is not a square and has long and short sides, the difference in bounding box shape and scale in the regression sample will lead to differences in its IoU value when the deviation and shape deviation are the same and not all 0. 
\par$\bullet$For bounding box regression samples of the same scale, when the deviation and shape deviation of the regression sample are the same and not all 0, the shape of the bounding box will have an impact on the IoU value of the regression sample. The change in IoU value corresponding to the deviation and shape deviation along the short edge direction of the bounding box is more significant.
\par$\bullet$For regression samples with the same shape bounding box, when the regression sample deviation and shape deviation are the same and not all 0, compared to larger scale regression samples, the IoU value of smaller scale bounding box regression samples is more significantly affected by the shape of the GT box
\subsection{Shape-IoU}
The formula for shape-iou can be derived from Fig.\ref{fig_3}:
\begin{equation}
	IoU = \frac{|B \cap B^{gt}|}{|B \cup B^{gt}|}
	\end{equation}
	\begin{equation}
		ww = \frac{2 \times (w^{gt})^{scale}}{(w^{gt})^{scale} + (h^{gt})^{scale}}
	\end{equation}
	\begin{equation}
	hh = \frac{2 \times (h^{gt})^{scale}}{(w^{gt})^{scale} + (h^{gt})^{scale}}
	\end{equation}
	\begin{equation}
	{distance}^{shape} = hh \times (x_c - x_c^{gt})^2/c^2 + ww \times (y_c - y_c^{gt})^2/c^2
	\end{equation}
	\begin{equation}
		\Omega^{shape} = \sum_{t=w,h} (1 - e^{-\omega_t})^\theta,\theta=4
	\end{equation}
	\begin{equation}
		\begin{cases}
			\omega_w = hh \times \frac{|w - w^{gt}|}{\max(w, w^{gt})} \\
			\omega_h = ww \times \frac{|h - h^{gt}|}{\max(h, h^{gt})}
		\end{cases}
	\end{equation}
	Where scale is the scale factor, which is related to the scale of the target in the dataset, and $ww$ and $hh$ are the weight coefficients in the horizontal and vertical directions respectively, whose values are related to the shape of the GT box. Its corresponding bounding box regression loss is as follows:
	\begin{equation}
		L_{\text{Shape-IoU}} = 1 - \text{IoU} + \text{distance}^{shape} + 0.5 \times \Omega^{shape}
	\end{equation}
\subsection{Shape-IoU in Small Target}
\subsubsection{Shape-Dot Distance}
We integrate the idea of Shape-IoU into Dot Distance to obtain Shape-Dot Distance, which is defined as follows:
	\begin{equation}
	D = \sqrt{hh \times (x_c - x_c^{gt})^2 + ww \times (y_c - y_c^{gt})^2}
\end{equation}
	\begin{equation}
S = \sqrt{\frac{\sum_{i=1}^M \sum_{j=1}^{N_i} w_{ij} \times h_{ij}}{\sum_{i=1}^M N_i}}
\end{equation}
\begin{equation}
{DotD}^{shape} = e^{-\frac{D}{S}}
\end{equation}
\subsubsection{Shape-NWD}
Similarly, we integrate the idea of Shape-IoU into NWD to obtain Shape-NWD, which is defined as follows:
\begin{equation}
	B =  \frac{(w - w_{gt})^2 + (h - h_{gt})^2}{{weight}^2},weight=2
\end{equation}
\begin{equation}
D = \sqrt{hh \times (x_c - x_c^{gt})^2 + ww \times (y_c - y_c^{gt})^2 + B}
\end{equation}
\begin{equation}
{NWD}^{shape} = e^{-\frac{D}{C}}
\end{equation}
\section{Experiments}
\subsection{PASCAL VOC on YOLOv8 and YOLOv7}
The PASCAL VOC dataset is one of the most popular datasets in the field of object detection, in this article we use the train and val of VOC2007 and VOC2012 as the training set including 16551 images, and the test of VOC2007 as the test set containing 4952 images. In this experiment, we choose the state-of-the-art one-stage detector YOLOv8s and YOLOv7-tiny to perform comparison experiments on the VOC dataset, and SIoU is chosen as the comparison method for the experiments. The experimental results are shown in TABLE\ref{tab:mytable1}:
\begin{table}[h]
	\centering
		\begin{tabular}{ccc}
			\toprule 
		& $AP_{50}$ & $mAP_{50:95}$ \\
		\midrule 
		Yolov7+SIoU & 63.4 & 37.3 \\
		Yolov7+Shape-IoU & 63.9(+0.5) & 37.9(+0.6) \\
			\midrule 
		Yolov8+SIoU & 69.5 & 48.3 \\
		Yolov8+Shape-IoU & 70.1(+0.6) & 48.8(+0.5) \\
		\bottomrule 
		\end{tabular}
		\caption{The performance of SIoU and Shape-IoU losses on Yolov7 and Yolov8.}
	\label{tab:mytable1}
\end{table}
\subsection{VisDrone2019 on YOLOv8}
VisDrone2019 is the most popular UAV aerial imagery dataset in the field of object detection, which contains a large number of small targets compared to the general dataset. In this experiment, YOLOv8s is chosen as the detector and the comparison method is SIoU. The experimental results are as follows:
\begin{table}[h]
	\centering
	\begin{tabular}{ccc}
		\toprule 
		& $AP_{50}$ & $mAP_{50:95}$ \\
		\midrule 
		Yolov8+SIoU & 38.8 & 22.6 \\
		Yolov8+Shape-IoU & 39.1(+0.3) & 22.8(+0.2) \\
		\bottomrule 
	\end{tabular}
	\caption{The performance of SIoU and Shape-IoU losses on Yolov8.}
	\label{tab:mytable2}
\end{table}
\subsection{AI-TOD on YOLOv5}
AI-TOD is a remote sensing image dataset, which is different from general datasets in that it contains a significant amount of tiny targets, and the average size of the targets is only 12.8 pixels. In this experiment, YOLOv5s is chosen as the detector, and the comparison method is SIoU. The experimental results are shown in TABLE\ref{tab:mytable3}:
\begin{table}[h]
	\centering
	\begin{tabular}{ccc}
		\toprule 
		& $AP_{50}$ & $mAP_{50:95}$ \\
		\midrule 
		Yolov5+SIoU & 42.7 & 18.1 \\
		Yolov5+Shape-IoU & 44.3(+1.6) & 18.7(+0.6) \\
		\bottomrule 
	\end{tabular}
	\caption{The performance of SIoU and Shape-IoU losses on Yolov5.}
	\label{tab:mytable3}
\end{table}
\section{Conclusion}
\par In this article, we summarize the benefits and disadvantages of existing bounding box regression methods, pointing out that existing research methods focus on considering the geometric constraints between the GT box and the predicted box, while ignoring the influence of geometric factors such as the shape and scale of the bounding box itself on the regression results. Then, by analyzing the regression characteristics of the bounding boxes, we discovered rules in which the geometric factors of the bounding boxes themselves can affect the regression. Based on the above analysis, we propose the Shape-IoU method, which can focus on the shape and scale of the bounding box itself to calculate losses, thereby improving accuracy. Finally, a series of comparative experiments were conducted using the most advanced one-stage detector on datasets of different scales, and the experimental results showed that our method outperformed existing methods and achieved state-of-the-art performance.

\end{document}